%% file: Paper-3940.tex
\documentclass[runningheads]{llncs}
\usepackage[T1]{fontenc}
\usepackage{graphicx,verbatim}
\usepackage{float}
\usepackage[T1]{fontenc}
\usepackage{booktabs}
\usepackage{amsmath}
\usepackage{amssymb}
\usepackage{graphicx}
\usepackage{multicol}
\usepackage{multirow}
\usepackage{adjustbox}
\usepackage{hyperref}
\usepackage{multicol}
\usepackage{multirow}
\usepackage{makecell}
\usepackage{pifont}
\hypersetup{
    colorlinks=true,
    linkcolor=blue,
    filecolor=magenta,      
    urlcolor=cyan,
    pdftitle={3D Graph Diffusion},
    pdfpagemode=FullScreen,
    }
\usepackage{orcidlink}
\usepackage{url}

\begin{document}
\title{Semantically Consistent Discrete Diffusion for 3D Biological Graph Modeling}
\titlerunning{Semantically Consistent Discrete Diffusion}
\author{
    Chinmay~Prabhakar\thanks{Contributed equally} \inst{1,2} \orcidlink{0000-0002-1780-8108}\and
    Suprosanna~Shit$^\star$ \inst{1,2} \orcidlink{0000-0003-4435-7207} \and
    Tamaz~Amiranashvili \inst{1, 3}\orcidlink{0000-0001-8914-3427} \and
    Hongwei~Bran~Li\inst{4} \orcidlink{0000-0002-5328-6407}\and
  Bjoern~Menze\inst{1} \orcidlink{0000-0003-4136-5690}}

\authorrunning{Prabhakar and Shit et al.}
\institute{Department of Quantitative Biomedicine, University of Zurich, Switzerland \and
ETH AI Center, ETH Zurich, Switzerland \and
 Department of Computer Science, Technical University of Munich, Germany \and
 Athinoula A. Martinos Center, Harvard Medical School, USA
\email{chinmay.prabhakar@uzh.ch}}
   
\maketitle              
\begin{abstract}
3D spatial graphs play a crucial role in biological and clinical research by modeling anatomical networks such as blood vessels, neurons, and airways. However, generating 3D biological graphs while maintaining anatomical validity remains challenging, a key limitation of existing diffusion-based methods. In this work, we propose a novel 3D biological graph generation method that adheres to structural and semantic plausibility conditions. We achieve this by using a novel projection operator during sampling that stochastically fixes inconsistencies. Further, we adopt a superior edge-deletion-based noising procedure suitable for sparse biological graphs. Our method demonstrates superior performance on two real-world datasets, human circle of Willis and lung airways, compared to previous approaches. Importantly, we demonstrate that the generated samples significantly enhance downstream graph labeling performance. Furthermore, we show that our generative model is a reasonable out-of-the-box link predictior.

\end{abstract}
\input{01_intro}
\input{02_rel_lit}
\input{03_method}
\input{04_exp}
\section{Conclusion}
Maintaining semantic validity is a significant obstacle to adopting diffusion-based biological graph generation methods. In this work, we ensure that generated graphs adhere to both structural and edge-label constraints. To the best of our knowledge, this is the first attempt to correct potential inconsistencies in generated samples stochastically. It makes our method particularly attractive to scenarios where a combination of multiple constraints reduces the set of acceptable outputs, rendering a naive rejection strategy suboptimal. Most importantly, our method addresses the major limitation of current data-driven biological graph generation models and makes discrete diffusion models more practical. 

\begin{credits}
\subsubsection{\ackname} This work has been supported by the Helmut Horten Foundation. S. S. is supported by the UZH Postdoc Grant (K-74851-03-01). H. B. Li is supported by an SNF postdoctoral mobility grant.

\subsubsection{\discintname}
The authors declare no competing interests for this article.
\end{credits}

\bibliographystyle{splncs04}
\bibliography{Paper-3940}
\end{document}

%% file: 01_intro.tex
\section{Introduction}
\label{sec:intro}

3D spatial graphs are an important representation for the study of biological graphs, including vascular networks \cite{paetzold2021whole}, neural connectivity \cite{yang2024morphgrower}, airway trees~\cite{nousias2020avatree}, lymphatic networks \cite{savinkov2020graph}, and bone microarchitecture \cite{pothuaud2000new}. These graphs are obtained from the segmentation \cite{wittmann2025vesselfm} of imaging modalities. Several works \cite{rahaghi2016pulmonary,van2019airway} have studied the characteristics of each of these graphs and their relevance as biomarkers for detecting abnormalities. Learning the distribution of biological graphs is crucial not only because it can serve as a way to generate synthetic samples but also because it can be used as a strong prior for downstream tasks such as link prediction, labeling, etc. However, learning the distribution of biological graphs is challenging due to the difficulty of enforcing anatomical plausibility, both in terms of structural constraints (e.g., avoiding cycles in airway trees) and semantic coherence (e.g., preventing anatomically inconsistent edge labels between neighboring nodes).

Biological graph modeling has been studied using stochastic methods \cite{rauch2021interactive} or topology optimization \cite{jessen2022rigorous}. These models do not explicitly learn the underlying distribution but are rather sampled from a heuristic model. Recently, generative models \cite{prabhakar20243d} have been used to model vascular graphs. However, there is no guarantee of generating anatomically plausible graphs, often leading to reliance on sub-optimal post-processing or inefficient sample rejection. Further, the discrete noising model considered in \cite{prabhakar20243d} changes the ground truth edge class arbitrarily to a different edge class. This noising procedure is inefficient since the diffusion model wastes resources on discarding a significant amount of implausible long-range edges.
Recently, Madeira et al.~ \cite{madeira2025generative} introduced an alternative edge noising model that randomly deletes existing edges, which mitigates the problem of the appearance of implausible edges in the noising process.
Furthermore, they provided theoretical results showing that if a desired graph property remains invariant under a noising procedure, it can be enforced by a hard projection operator during sampling.
This allows one to generate graphs that maintain structural properties, such as a tree-like topology.
However, biological graphs often adhere to strict \emph{semantic} constraints on top of structural ones.
For instance, a basal artery (BA) must not be directly connected to the anterior cerebral artery (ACA).
This represents an important biological prior, which cannot be imposed by projections in~\cite{madeira2025generative}.
In this work, we go beyond structural constraints and introduce a projector module that stochastically corrects the intermediate graph by removing edges that violate structural constraints (e.g., causing cycles in the airway tree) \emph{and/or} semantic constraints (e.g., BA and ACA are direct neighbors).
In summary, our contributions are threefold:

\begin{enumerate}
    \item  We propose a novel label-consistent projector that, during sampling, prevents conflicting neighboring edge labels that are anatomically invalid, thus ensuring anatomically plausible graphs- as demonstrated in our experiments in two real-world datasets, human circle of Willis and lung airway.

 \item  We adopt an edge deletion noising model for the discrete diffusion, which is suitable for sparse biological graphs.

 \item  We validate the efficacy of generated samples from our proposed generation to train a vessel labeler that significantly boosts its accuracy. Furthermore, we demonstrate excellent out-of-the-box emerging link prediction properties of our graph generator in the airway dataset. Code is available at \footnote{\url{https://github.com/chinmay5/semantically_consistent_graph_generation}}

\end{enumerate}

%% file: 02_rel_lit.tex
\section{Related Literature}
\label{sec:rel_lit}
\paragraph{\textbf{Biological graph generation}} is a complex modeling task due to anatomical variations, making specialized heuristics-driven techniques very popular. For instance, stochastic generation \cite{reichold2009vascular,rauch2021interactive} and topology optimization \cite{jessen2022rigorous} rely on heuristics to generate realistic vessel trees, while Nousias et al. \cite{nousias2020avatree} applied heuristics to generate airway trees. Recently, data-driven methods have been proposed \cite{kuipers2024generating,prabhakar20243d,yang2024morphgrower,nousias2020avatree} to avoid such heuristics. Among these, Prabhakar and Shit et al. \cite{prabhakar20243d} propose a model capable of generating complex vascular graphs. Their method is extendable to other spatial structures. However, it can generate semantically invalid graphs, leading to data inconsistencies.

\paragraph{\textbf{Discrete diffusion models}} excel in modeling discrete data and are widely used in molecule generation \cite{vignac2022digress,vignac2023midi}, protein structure design \cite{gruver2023protein}, and language modeling \cite{austin2021structured,lou2023discrete}. However, they often fail to enforce domain-specific constraints, leading to invalid samples. To address this, Sharma et al. \cite{sharma2024diffuse} introduce plug-and-play closed-form guidance for properties such as edge count, degree, and triangle count, etc., but it does not guarantee constraint satisfaction. Recently, Madeira et al.~\cite{madeira2025generative} proposes hard-projection criteria to enforce validity. Note that their method does not consider edge labels and focuses only on structural constraints (tree structure, node-degree, etc).

%% file: 03_method.tex
\section{Methodology}
\label{sec:meth}
\begin{figure}[t!]
    \centering
    \includegraphics[width=0.95\textwidth, trim=20 05 20 13, clip]{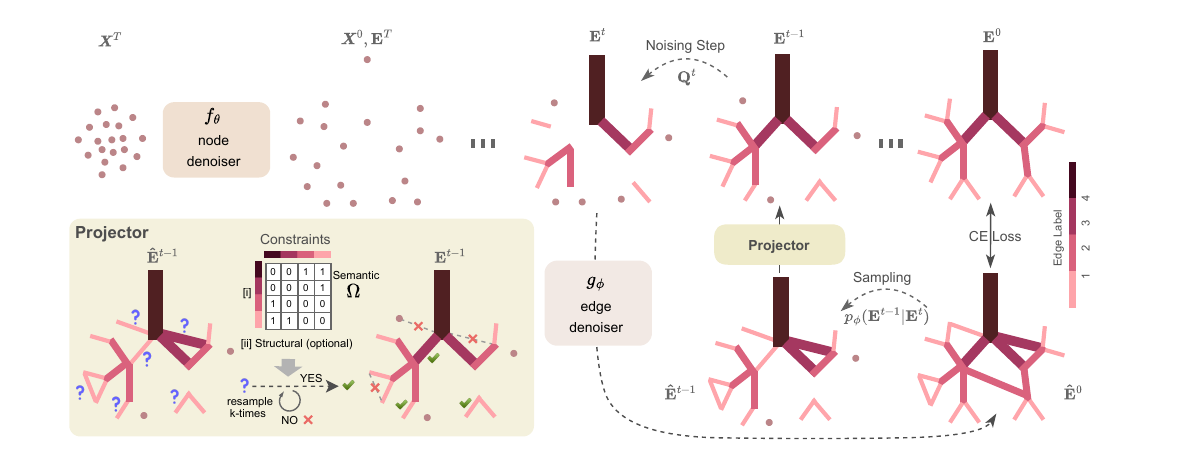}
    \caption{Our method first generates node coordinates using a point cloud denoiser. Subsequently, the edge denoiser is trained with an edge-deletion noise model. During inference, the edge denoiser suggests a set of new edges. Our novel semantic consistent projector discards an edge if none of $k$ resampled edge labels meet the validity criteria.}
    \label{fig:overview}
\end{figure}

\paragraph{\textbf{Problem Statement:}}
Our goal is to learn the distribution $p(G)$ of 3D spatial graphs $G:=(\boldsymbol{X},\mathbf{E})$, where $\boldsymbol{X} \in \mathbb{R}^{n \times 3}$ represents coordinates of $n$ nodes, $\mathbf{E} \in \mathbb{R}^{n \times n \times c}$ is adjacency matrix containing one-hot encoding of categorical edge labels indicating vessel class and $E$ is the corresponding set of edges. Crucially, we are interested in graphs that satisfy semantic (edge-label) consistency criteria $\boldsymbol{\Omega}\in \{0,1\}^{c \times c}$, where $\boldsymbol{\Omega}[c_i, c_j] =0$ indicates edge types $c_i$ and $c_j$ can be adjacent, and $1$ otherwise.
\noindent A graph $G$ satisfies $\boldsymbol{\Omega}$-constraint, i.e.:
\begin{equation}
  P_\Omega(E)=True, \mbox{ iff } \sum\limits_{v=1}^n \sum\limits_{\substack{i,j \in \mathcal{N}(v) \\ i \neq j}} \left\|\Big(\mathbf{E}[i,v] \otimes \mathbf{E}[v,j]^{\top}\Big) \odot\boldsymbol{\Omega}\right\|_F = 0 
\end{equation}
where $\mathcal{N}(v)$ is the set of neighbors of node $v$, $\otimes$ denotes tensor product, and $\odot$ denotes Hadamard product. Additionally, any available structural constraints can also be incorporated, e.g., tree structure for airways.

\paragraph{\textbf{Overview:}}
We adopt a two-stage graph generation strategy \cite{prabhakar20243d}, which is described in Fig. \ref{fig:overview}. In the first stage, a denoising diffusion probabilistic model \cite{ho2020denoising} is trained on the node coordinates. Specifically, Gaussian noise is added to coordinates $\boldsymbol{X}^0$ through the forward process:
$\boldsymbol{X}^{t} := \sqrt{\alpha^t_x} \boldsymbol{X}^{t-1} + \sqrt{1 - \alpha^t_x} \boldsymbol{\epsilon}, \quad \boldsymbol{\epsilon} \sim \mathcal{N}(\boldsymbol{0}, \boldsymbol{I}),
$ where $\alpha^t_x$ transitions from $1$ to $0$ with increasing $t$. A neural network $f_\theta$ predicts the noise $\boldsymbol{\hat{\epsilon}} := f_\theta(\boldsymbol{X}^t, t)$ using $\boldsymbol{X}^t := \sqrt{\bar{\alpha}^t_x} \boldsymbol{X}^0 + \sqrt{1 - \bar{\alpha}^t_x} \boldsymbol{\epsilon}$; ($\bar{\alpha}^t_x := \prod_{s=1}^t \alpha^s_x$). The model is trained using $\mathcal{L}_{\theta} := \mathbb{E}_{t, \boldsymbol{X}^0, \boldsymbol{\epsilon}} \left\|\boldsymbol{\epsilon}-\boldsymbol{\hat{\epsilon}}\right\|^2$. During sampling, node coordinates are denoised iteratively starting from $\boldsymbol{X}^T,\boldsymbol{\epsilon}_b \sim \mathcal{N}(\boldsymbol{0}, \boldsymbol{I})$:
$$\boldsymbol{X}^{t-1} = \dfrac{1}{\sqrt{\alpha^t_x}}\left(\boldsymbol{X}^t - \dfrac{1 - \alpha^t_x}{\sqrt{1 - \bar{\alpha}^t_x}} \boldsymbol{\hat{\epsilon}}^t\right) + \sqrt{1 - \alpha^t_x} \boldsymbol{\epsilon}_b.$$

In the second stage, a discrete diffusion process \cite{vignac2023midi,prabhakar20243d} is used on the edges, where a transition matrix $\boldsymbol{Q}^t := \alpha^t_e \boldsymbol{I} + (1 - \alpha^t_e) \mathbf{1}_c \boldsymbol{m}^\top$ changes the edges independently during forward noising as $\mathbf{E}^t := \mathbf{E}^{t-1} \boldsymbol{Q}^t$. A non-rotation-equivariant graph transformer $g_\phi$ is trained to predict $\mathbf{\hat{E}}^0 := g_\phi(\mathbf{E}^t, \boldsymbol{X}^0, t)$. The model is trained using cross-entropy loss using
$\mathcal{L}_{\phi} := \mathbb{E}_{t, \mathbf{E}^0} \sum_{ij} CE(\mathbf{E}^0_{ij},\mathbf{\hat{E}}^0_{ij})$.
During sampling, we sample edges from the posterior:
$$p_\phi(\mathbf{E}^{t-1}_{ij} | \mathbf{E}^t_{ij}) := \dfrac{\mathbf{E}^t_{ij} {\boldsymbol{Q}^t}^{\top} \odot \mathbf{\hat{E}}^0_{ij} \bar{\boldsymbol{Q}}^{t-1}_{ij}}{\mathbf{\hat{E}}^0_{ij} \bar{\boldsymbol{Q}}^t_{ij} {\mathbf{E}^t}^{\top}_{ij}},$$
starting from $\mathbf{E}_{ij}^T \sim \boldsymbol{m}$, where $\bar{\boldsymbol{Q}}^t:=\boldsymbol{Q}^1 \ldots \boldsymbol{Q}^t$. Note that Prabhakar and Shit et al.~\cite{prabhakar20243d} use uniform $\boldsymbol{m}$ and can not guarantee structural and semantic constraints; thus, it is prone to generating \emph{semantically invalid} graphs.

\paragraph{\textbf{Edge-deletion-based Noising:}} 
Discrete denoising diffusion models typically rely on uniform or marginal noise to perturb graph connectivity \cite{vignac2023midi,vignac2022digress,prabhakar20243d}. However, real-world spatial graphs are often sparse (i.e., the number of edges grows linearly with the number of nodes), so standard noise schedulers introduce excessive, implausible connections early on. This contradicts the assumption that samples should remain near the data manifold during early diffusion steps, leading to inefficiency. Recently, \cite{lou2023discrete,austin2021structured} showed that discrete diffusion models benefit from an \emph{absorbing state} noising process, mapping each token to a [MASK] in language modeling. Madeira et al.~\cite{madeira2025generative} adapted this to graphs by treating the absence of an edge as the [MASK] token, so each edge is either deleted or retains its existing label (never changing labels).  The transition matrix for noising step is $ \boldsymbol{Q}^t:= \alpha^t_e \boldsymbol{I} + (1 - \alpha^t_e)\vec{1}_c \boldsymbol{e}_E^{\top}$, where $\boldsymbol{e}_E$ is the one hot encoding of the absence of edge. At $t=T$, all the edges are deleted, and we get an empty graph with only node coordinates. During sampling, only new edges are inserted, and once inserted, they remain fixed throughout denoising.

\paragraph{\textbf{Semantically Consistent Projection:}}
During edge denoising at time step $t$, $\mathbf{\hat{E}}^{t-1}$ is sampled from categorical distribution $p_\phi(\mathbf{E}^{t-1} | \mathbf{E}^t)$. Since the forward noising is designed by edge deletion, during denoising, the network learns to add edges. There is no guarantee that $\mathbf{\hat{E}}^{t-1}$ will satisfy structural or semantic constraints. Some newly added edges can violate structure and/or semantic constraints, creating an invalid intermediate graph. 
Recently, \cite{madeira2025generative} showed that if the edge-deletion noising process preserves a structural constraint, then projecting intermediate graphs during denoising onto a constrained set ensures compliance with that constraint.

This key result motivates us to investigate semantic properties relevant to spatial biological graphs that remain invariant under the edge-deletion noising model. Since the edge-deletion process does not change the relationships among neighboring edges, we can devise a label-aware projector module to fix the intermediate graph if needed. Importantly, existing work \cite{madeira2025generative} only considered structural properties, resulting in either accepting or rejecting new edges. This is useful to prevent the formation of cycles in many biological graphs, e.g., airways. However, in the case of multi-class edges, the model can propose a structurally consistent edge that only violates semantic constraints. A naive strategy of simply deleting these edges is suboptimal, as it penalizes the existence of an edge based on one sampled semantic label, resulting in overly sparse graphs.

To this end, we propose a novel label-aware projector, which allows us to sample from the posterior distribution constrained by our semantic plausibility. The projector begins with $E^{t-1}\leftarrow E^{t}$ and adds edges in $E^{t-1}$ that satisfy the semantic-consistency criteria. Specifically, for each edge $e_{ij}\in  \hat{E}^{t-1}\setminus  E^{t}$ our projector adds $E^{t-1} \leftarrow E^{t-1} \cup \{e_{ij}\}$ if $P_\Omega(E^{t-1}\cup \{e_{ij}\})=True$. If not, we sample $k$ new edge labels from $e_{ij}\sim p_\phi(\mathbf{E}^{t-1}_{ij} | \mathbf{E}^t_{ij})$. Note that this is in contrast with \cite{madeira2025generative}, which either accept or discard edges, whereas we harness stochastic variation in favor of increased new edge proposals. The projector accepts the edge if any of the k resampled edge labels satisfy the semantic consistency; otherwise, it rejects the edge. This strategy is visually demonstrated in Fig. \ref{fig:overview}. Such a strategy allows the model to fix invalid edges stochastically and avoids overly sparse graphs, leading to semantically valid outputs. Crucially, our semantic projector can be seamlessly combined with a structural projector.

%% file: 04_exp.tex
\section{Experiments}
\label{sec:exp}

\paragraph{\textbf{Datasets:}} We use two publicly available datasets for the experiments. First is the circle of Willis (CoW) dataset consisting of 300 MRA from CROWN \cite{vos2024results} and 90 from TopCoW \cite{yang2023benchmarking} challenges. Please note that we use only CROWN samples to train the diffusion model. The TopCoW samples contain verified segmentations and are used to evaluate the vessel labeler.
The CROWN graphs generated following \cite{prabhakar20243d} contain 14 edge classes representing different artery labels.
The second dataset is from the Airway Tree Modeling (ATM) \cite{zhang2023multi} challenge, consisting of 300 CT images. We extracted the graph from the ground-truth segmentation using Voreen \cite{drees2021scalable}, then used radius information to define five thickness-based edge classes (including a background/no-edge class). We manually removed erroneous samples, yielding a final training set of 251 graphs. The number of nodes ranges between 40 and 500 in ATM (average 127) and between 13 and 27 in CoW (average 8). Such diversity underscores the generalizability of our method.

\paragraph{\textbf{Baselines:}} We use the two-stage diffusion model proposed by \cite{prabhakar20243d} as our baseline. We refer to it as Uniform based on its noising scheme. Additionally, we use Construct \cite{madeira2025generative}, which uses only structural constraints. We also include an ablation of our method, namely ours (\emph{k=0}), that deletes edges with label violations while our model resamples k (= 4) times to fix these violations.

\paragraph{\textbf{Metrics:}}
Following \cite{prabhakar20243d,feldman2023vesselvae,kuipers2024generating}, we evaluate by comparing various physical features of the generated samples, including node degree ({deg($\mathcal{V}$)}), number of edges (|$\mathcal{E}$|), length of edges ($l_\mathcal{E}$), and angle between the edges ($\mathcal{E}~\angle$). We also compute Betti-0 ($\beta_0$) for connected components and Betti-1 ($\beta_1$) for cycles. In addition, we report semantic validity (in \%) for the generated graphs. We report balanced accuracy (bal. acc.) and macro averaged F1 (avg. F1) scores for the vessel labeler and link prediction tasks. Additionally, we report semantic validity (in \%) for the latter.

\paragraph{\textbf{Semantic Validity (S.V.):}} We use edge-label based criterion for semantic validity. \textbf{CoW:} The CoW graph has no structural constraints, and the TopCoW annotation protocol \cite{yang2023benchmarking} determines its validity, please refer \cite{yang2023benchmarking} for details. For CoW, we have 13x13 $\Omega$ matrix for 13 edge-labels. \textbf{ATM:} An airway tree is invalid if (i) it contains a cycle, and (ii) it violates radius-based label hierarchy (e.g., connects trachea with bronchioles). We have 4x4 $\Omega$ as shown in  Fig. \ref{fig:overview}.

\begin{figure}[htbp]
    \centering
    \begin{tabular}{@{}%
        >{\centering\arraybackslash}p{0.12\linewidth} 
        >{\centering\arraybackslash}p{0.16\linewidth} 
        >{\centering\arraybackslash}p{0.15\linewidth} | 
        >{\centering\arraybackslash}p{0.16\linewidth} 
        >{\centering\arraybackslash}p{0.14\linewidth} 
        >{\centering\arraybackslash}p{0.16\linewidth}@{}}
         \mbox{Input} & \mbox{Prediction} & \mbox{Reference} & \mbox{Input} & \mbox{Prediction} & \mbox{Reference} \\ 
         \multicolumn{3}{c|}{\raisebox{-0.5\height}{\includegraphics[width=0.45\linewidth]{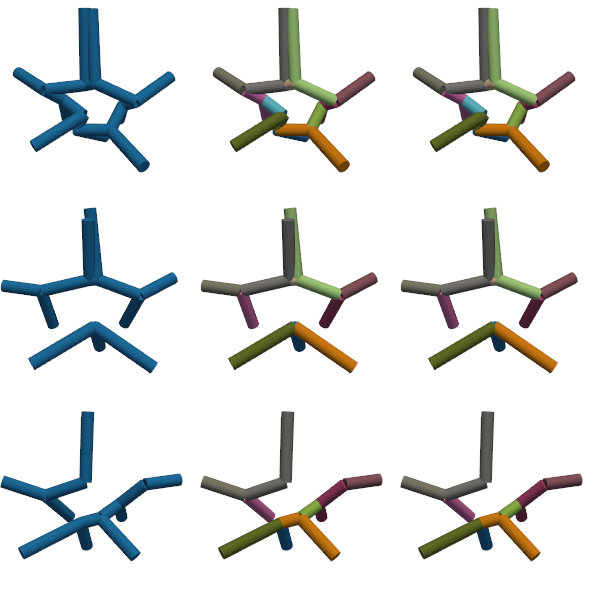}} \raisebox{-0.5\height}{\includegraphics[width=0.025\linewidth]{figures/color_legend.png}}} &
         \multicolumn{3}{c}{\raisebox{-0.5\height}{\includegraphics[width=0.45\linewidth]{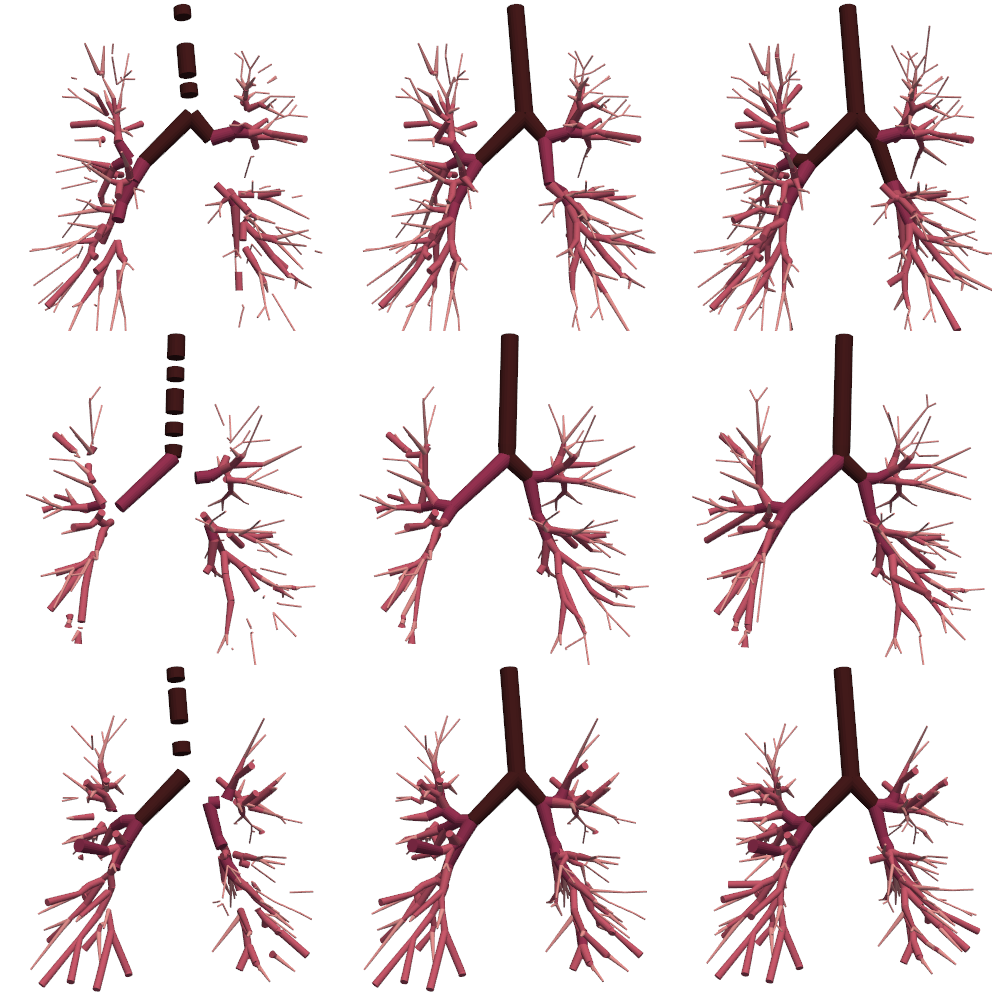}} \raisebox{-0.5\height}{\includegraphics[width=0.025\linewidth]{figures/color_legend1.png}}} 
    \end{tabular}
    \caption{Downstream applications of the diffusion model. \textbf{Left:} The vessel labeler receives circle of Willis graphs without any label and assigns multi-class labels based on TopCoW annotation protocol~\cite{yang2023benchmarking}. \textbf{Right:} The diffusion model predicts missing links on the airway tree samples. We show the ground truth samples for reference.}
    \label{fig:downstream}
\end{figure}

\paragraph{\textbf{Downstream Tasks:}}We asses our model on two downstream tasks (Fig. \ref{fig:downstream}). 

\noindent \textbf{(i) Vessel labeling} involves inferring multi-class labels to an unlabeled CoW graph based on the TopCoW annotation protocol \cite{yang2023benchmarking}. We generate 1000 synthetic samples to train the labeler and evaluate it on \emph{unseen} TopCoW samples. Specifically, we use a graph transformer to predict edge labels for the unlabeled edges and assess its performance against the reference labels. We use 30 samples as a test set and a 50/10 (train/val) split for the real-data baseline.

\noindent \textbf{(ii) Link prediction} involves predicting plausible connections on an ATM graph with approximately 30\%  missing edges. We feed this incomplete graph into our diffusion model for 100 denoising steps. Because our model only adds edges during denoising, it can predict missing links. In contrast, \cite{prabhakar20243d} relies on a uniform noising process and thus cannot be used for link prediction. To the best of our knowledge, this is the first attempt to use an out-of-the-box diffusion model for link prediction on biological graphs.

\paragraph{\textbf{Implementation Details:}}
We use the same node and edge denoising architecture as proposed by \cite{prabhakar20243d}. We use 8 edge-denoising blocks for CROWN and 5 for ATM dataset.
All models (including baselines) are trained from scratch for 1000 epochs on a single RTX A6000 GPU, with a batch size of 64 for CROWN and 4 for ATM with AdamW optimizer with a learning rate of 3e-4. The discrete-diffusion model uses a linear noise schedule and 500 noising steps.

\begin{table*}[t!]
  \caption[table: Comparison]{ \footnotesize Comparison of our method against other graph generation methods. Our proposed solution adheres to the data distribution statistics while satisfying the semantic validity (S.V.) constraints. KL statistics are reported in 10\textsuperscript{-3} and S.V. in \%.}
  \label{tab:results_SOTA}
  \centering
  \scriptsize
  \setlength{\tabcolsep}{2mm}{
   \begin{tabular}{l |l | c | c | c| c | c| c | c}
    \toprule
     & {Methods} &\textit{deg($\mathcal{V}$)} $\downarrow$ & |$\mathcal{E}$| $\downarrow$ &  $l_\mathcal{E}$ $\downarrow$ & $\mathcal{E} ~\angle \downarrow$& \textit{$\beta_0$}$\downarrow$ &\textit{$\beta_1$}$\downarrow$ & {S.V.\%}$\uparrow$\\
    \midrule
     \multirow{3}*{\rotatebox[origin=c]{90}{CROWN}} & Uniform \cite{prabhakar20243d} & 1.328 & 9.113 & 4.718 &9.550 & 3.284 & \textbf{1.069} & 70 \\
     
     & Construct \cite{madeira2025generative} & 1.017 & 5.867 & 0.862 & 5.178 & 1.510 & 2.833 & 75 \\
     
     & Ours (k=0) & 1.071 & 6.054 & \textbf{0.789} & 5.199 & 1.348 & 2.737 & \textbf{100} \\
    
    & \textbf{Ours} &\textbf{1.014} & \textbf{4.893}& 0.864 & \textbf{5.174} & \textbf{1.047} & 2.007 & \textbf{100} \\
    \midrule
    
     \multirow{3}*{\rotatebox[origin=c]{90}{ATM}} & Uniform \cite{prabhakar20243d} & 8.432 & \textbf{2.696} & \textbf{5.074} & 4.369 & \textbf{3.077} & 1.425 & 10\\
     
     & Construct \cite{madeira2025generative} & 2.301 & 2.734 & 6.089 & 4.350 & 9.065 & \textbf{0.0} & 40 \\
     
     & Ours (k=0) & 2.319 & 3.707 & 6.112 & 4.349 & 10.31 & \textbf{0.0} & \textbf{100} \\
     
     &\textbf{Ours} & \textbf{2.278} &  2.734 & 6.091 & \textbf{4.340} & 9.070 & \textbf{0.0} & \textbf{100}\\

    \bottomrule
  \end{tabular}
}
\end{table*}
\begin{figure}[htbp]
    \centering
    \begin{tabular}{@{}cc|c|c|cc@{}}
         & \mbox{Uniform} & \mbox{Construct} & \mbox{Ours} & \mbox{Ground Truth} &\\
         \rotatebox[origin=l]{90}{\mbox{~~~~~~CROWN}} & 
         \includegraphics[width=0.22\linewidth]{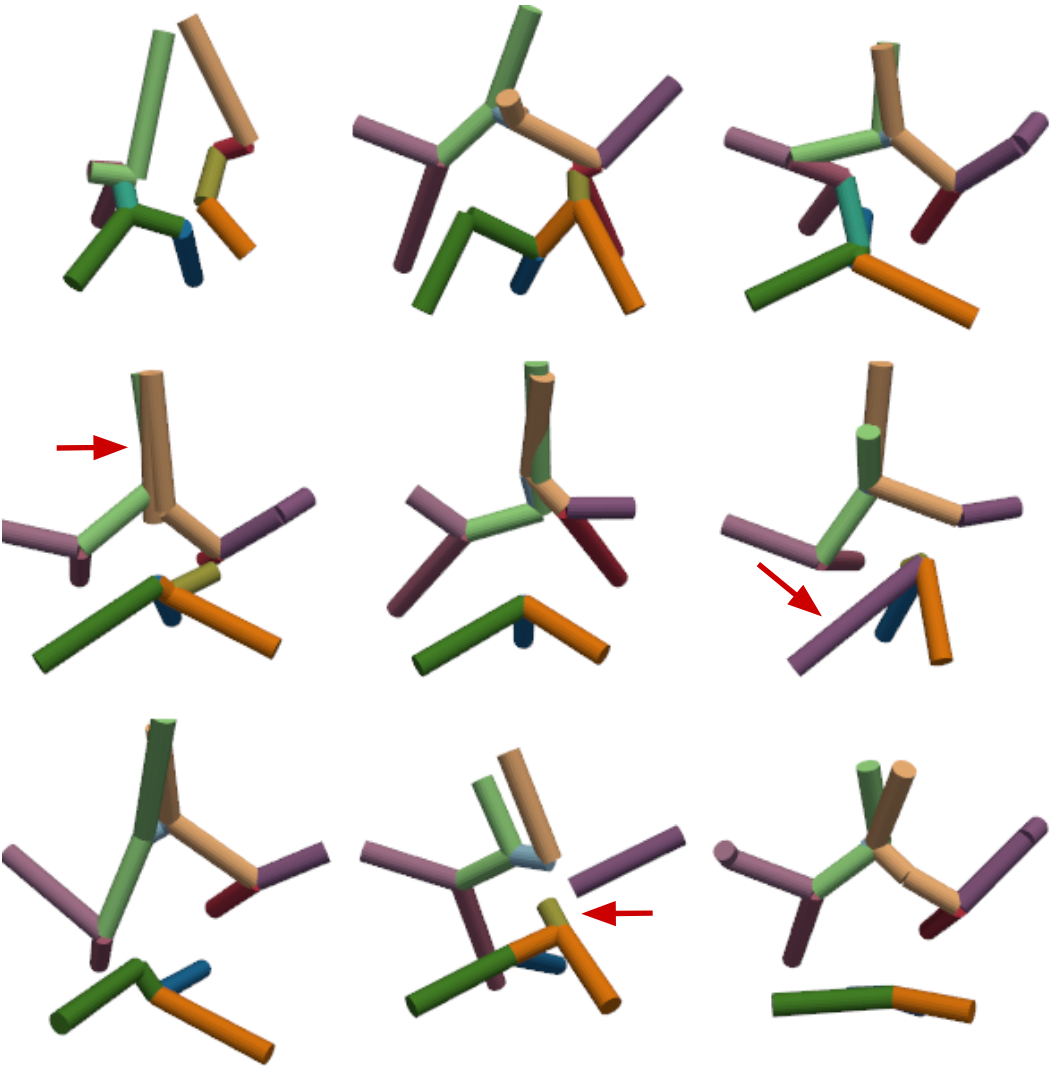} &
         \includegraphics[width=0.22\linewidth]{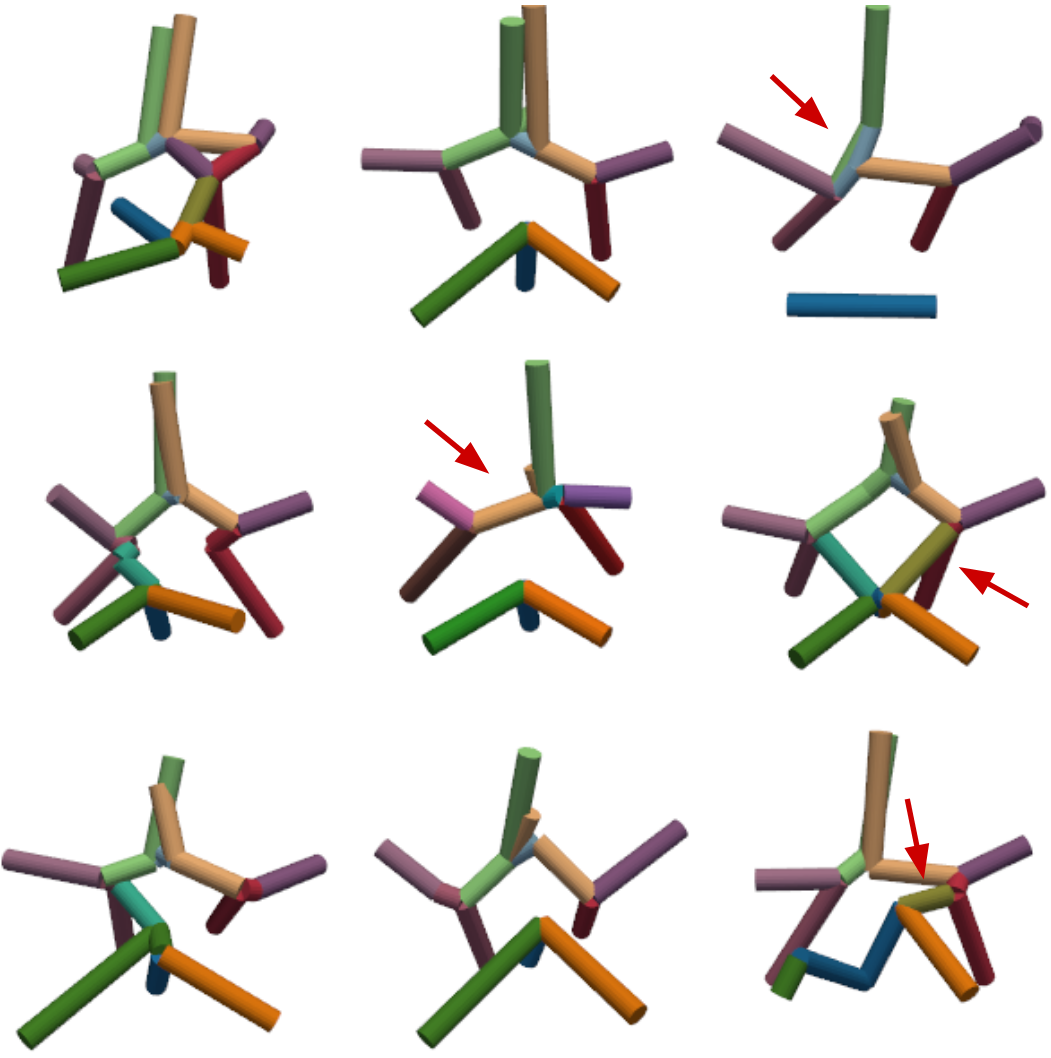} &
         \includegraphics[width=0.22\linewidth]{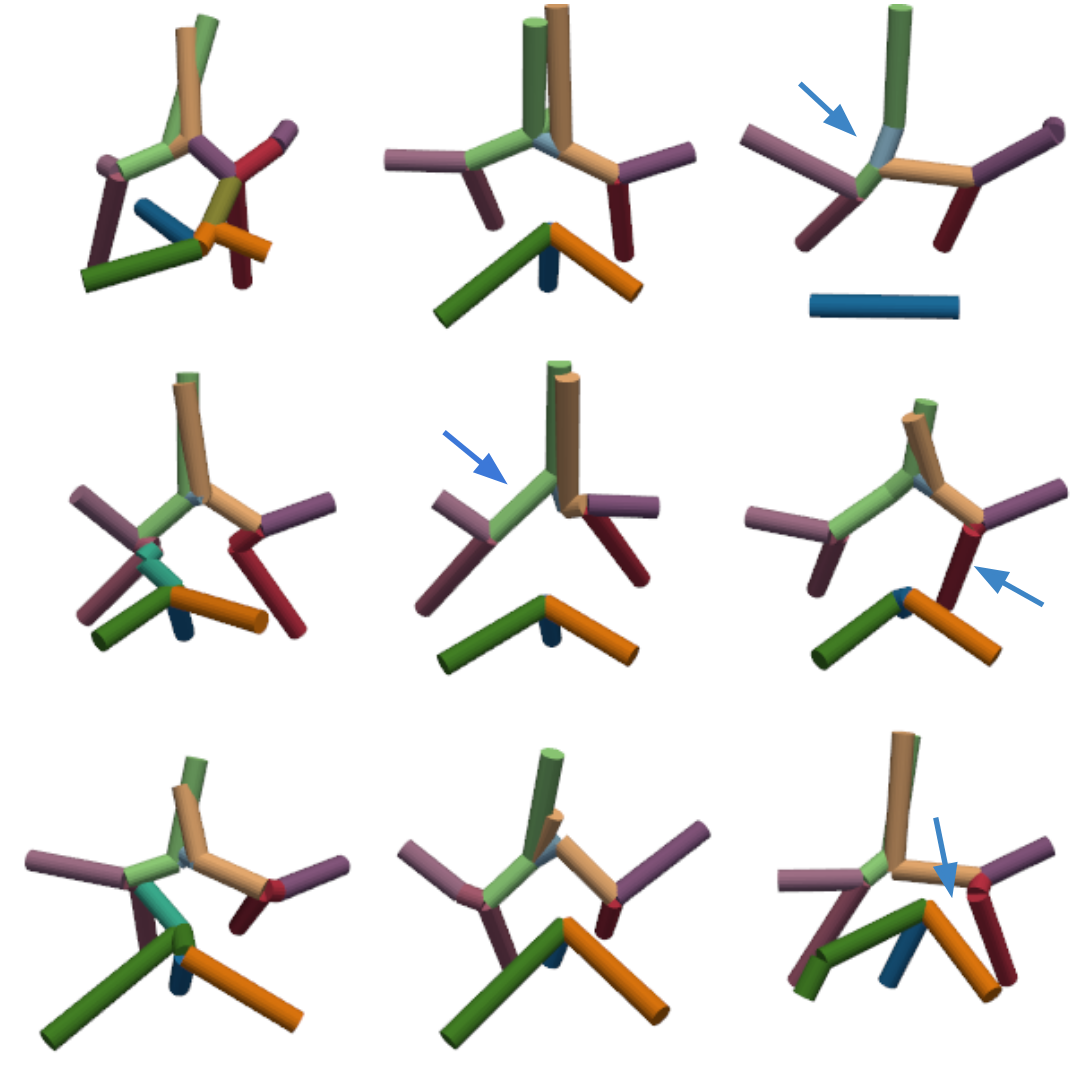} &
         \includegraphics[width=0.22\linewidth]{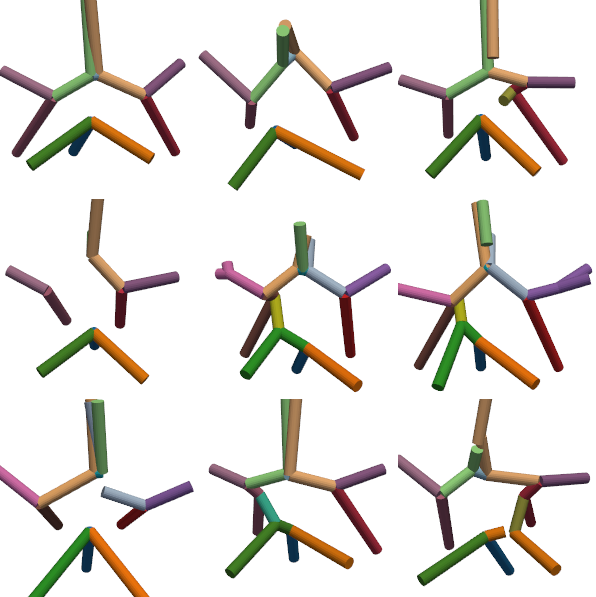} &  \includegraphics[width=0.018\linewidth]{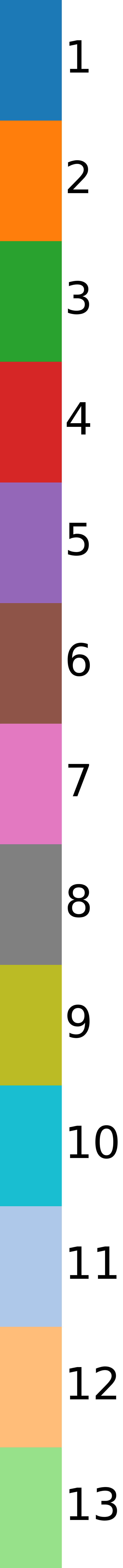}\\\hline
         \rotatebox{90}{\mbox{~~~~~~~~~~ATM}} & 
         \includegraphics[width=0.22\linewidth]{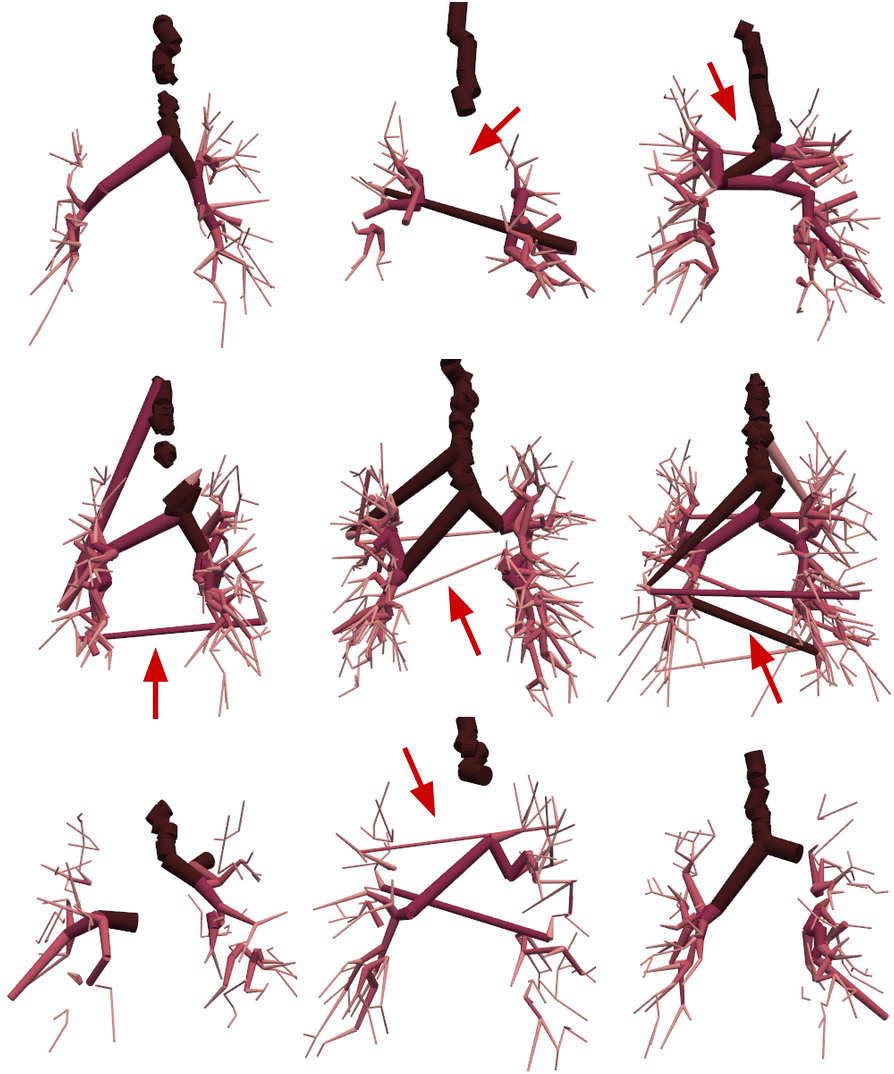} &
         \includegraphics[width=0.22\linewidth]{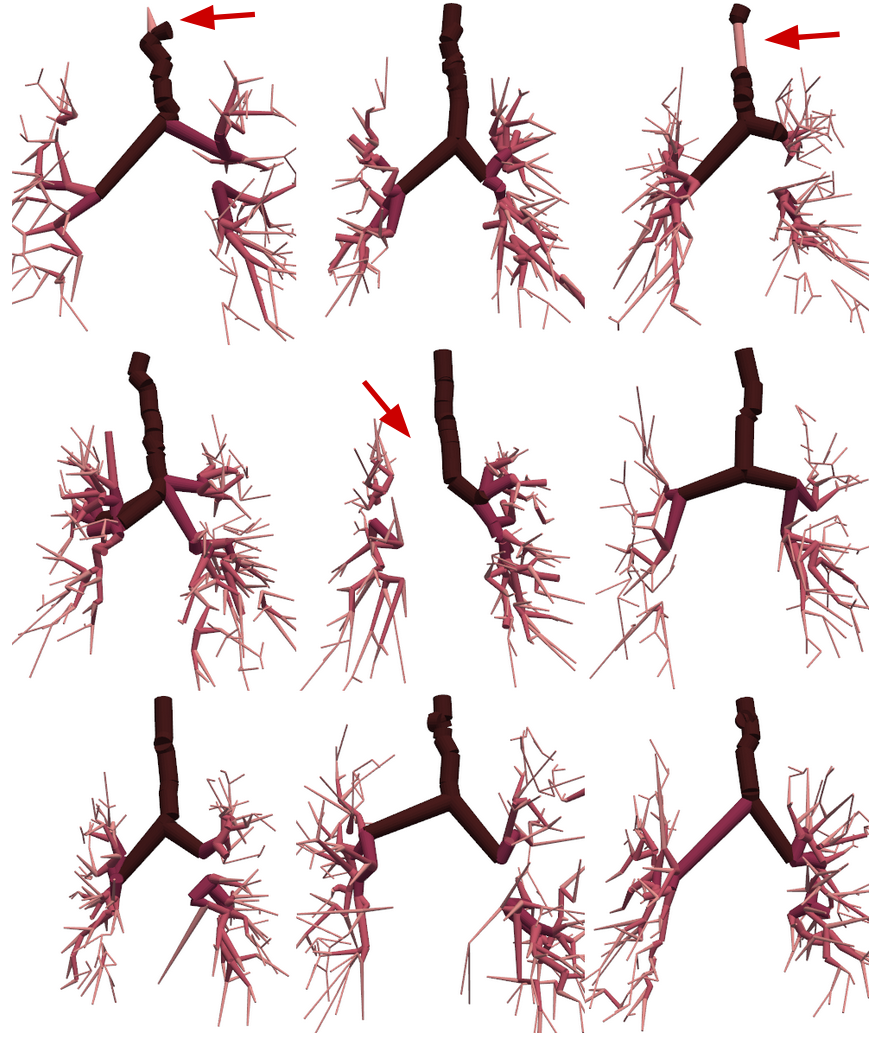} &
         \includegraphics[width=0.22\linewidth]{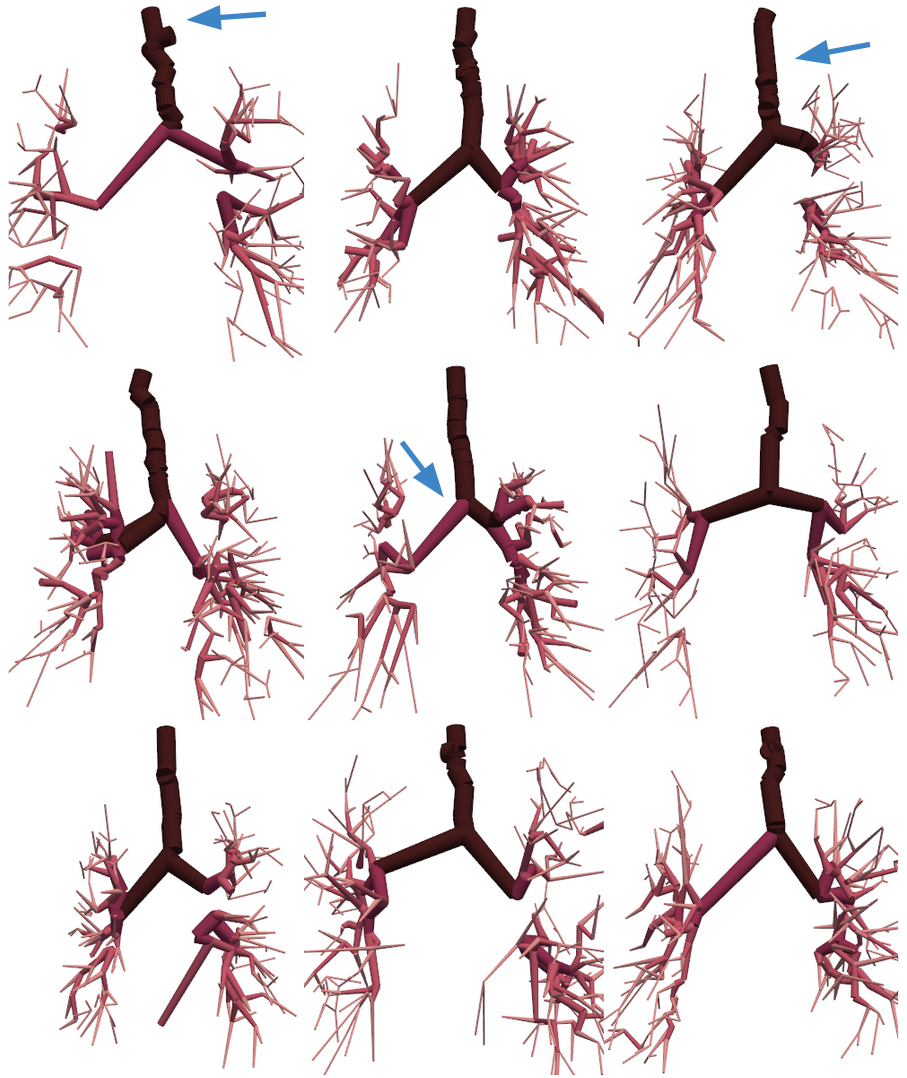} &
         \includegraphics[width=0.22\linewidth]{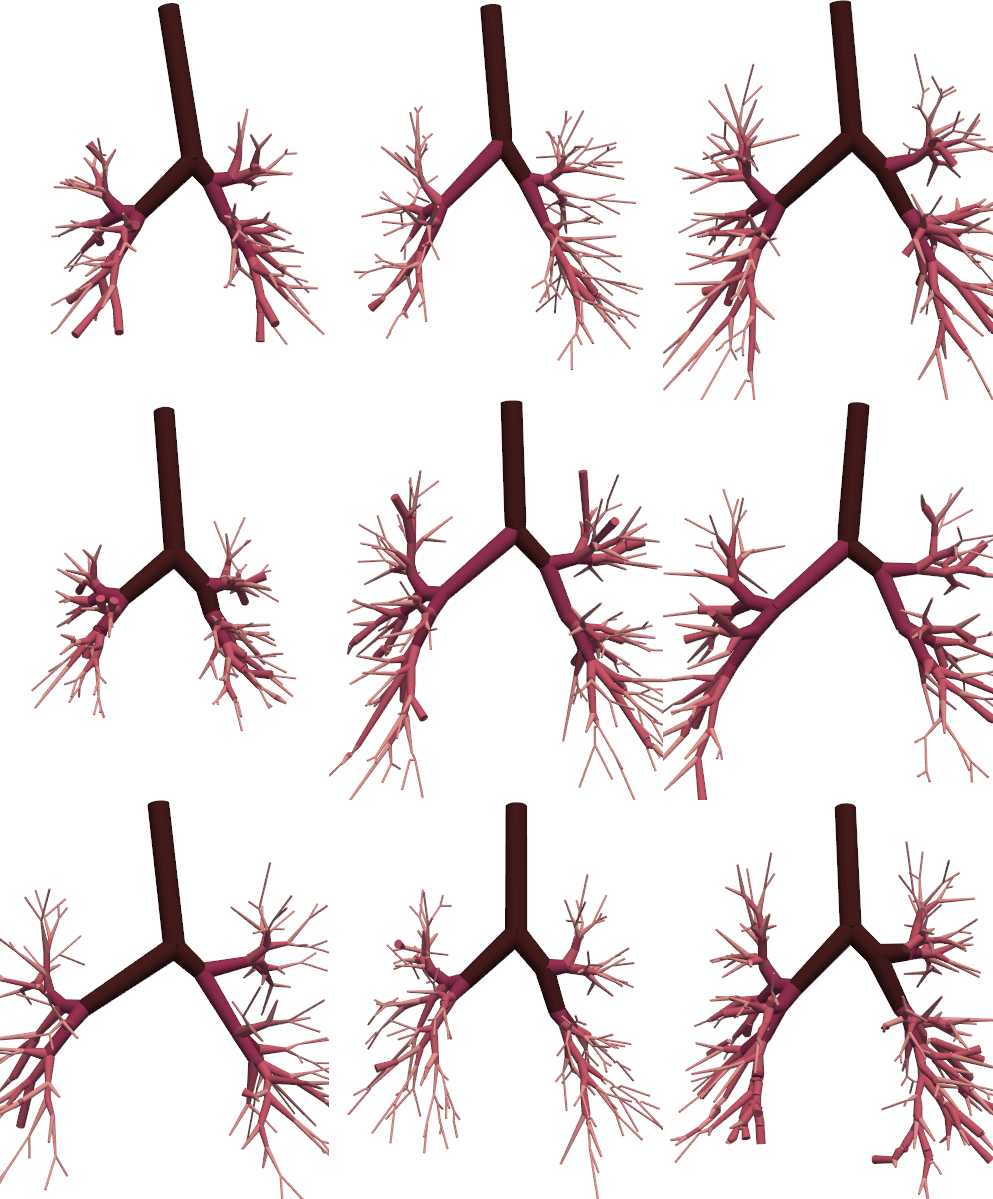}&  \includegraphics[width=0.02\linewidth]{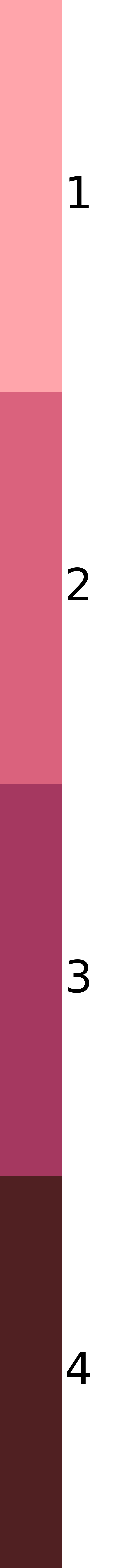}
    \end{tabular}
    \caption{\footnotesize Qualitative examples of graphs generated by \cite{prabhakar20243d}, Construct, and our method compared to ground truth samples. The constraint violations are highlighted in red.\cite{prabhakar20243d} produces samples with label and structural violations (cycles in airway trees). Construct can produce samples with label violations. Our method tries to adhere to both label and structural constraints, potentially fixing the issues (shown in blue).}
    \label{fig:qual}
\end{figure}

\begin{table*}[t!]
\centering
\begin{minipage}{0.45\linewidth}
\centering
\scriptsize
\caption{Vessel labeling on \textit{TopCoW}.}
\label{tab:topCoW_labeler}
\begin{tabular}{l c c}
\toprule
\textbf{Train Data} & \textbf{bal. acc.} $\uparrow$  & \textbf{avg. F1}  $\uparrow$ \\
\midrule
Uniform \cite{prabhakar20243d}    & 88.44   & 0.874 \\
Construct \cite{madeira2025generative} & 89.71   & 0.899 \\
Ours (k=0) & 94.68   & 0.961 \\
Ours  & \textbf{95.45} & \textbf{0.967} \\
\midrule
Real data & 93.77 & 0.946 \\
\bottomrule
\end{tabular}
\end{minipage}
\hfill
\begin{minipage}{0.52\linewidth}
\centering
\scriptsize
\caption{Link prediction on \textit{ATM}. Note that \cite{prabhakar20243d} can not be used for link prediction, while our method can predict the missing links.}
\label{tab:atm_labeler}
\begin{tabular}{l c c c}
\toprule
\textbf{Method} & \textbf{bal. acc.} $\uparrow$  & \textbf{avg. F1} $\uparrow$ & \textbf{S.V.} $\uparrow$ \\
\midrule
Construct \cite{madeira2025generative} & 84.32 & 0.842 & 70   \\
Ours (k=0)               & 84.83   & 0.849   & \textbf{100} \\
Ours & 84.97   & 0.849   & \textbf{100} \\

\bottomrule
\end{tabular}
\end{minipage}
\end{table*}

\paragraph{\textbf{Results on Circle of Willis:}}
\label{ssec:results}
In Table \ref{tab:results_SOTA}, which compares the statistics of the generated samples, we note that \cite{prabhakar20243d} performs well in structural metrics, but generates only 70\% semantically valid samples. Construct performs slightly better with 75\% semantic validity. Please note that CROWN has no structural constraints. Hence, the improved results compared to \cite{prabhakar20243d} highlight the advantage of the edge-deletion noising scheme over the uniform scheme; Ours(k=0) uses semantic constraints, and thus, it generates 100\% semantically-valid samples. However, deleting invalid edges makes the generated samples more sparse with worse distribution statistics. Our method leverages stochastic resampling and generates 100\% valid samples with better distribution compliance. The effect is further exemplified by the superior performance of the downstream vessel labeling task trained on the synthetic graphs (c.f. Tab.~\ref{tab:topCoW_labeler} and Fig. \ref{fig:downstream}).

\paragraph{\textbf{Results on lung airway:}} Table~\ref{tab:results_SOTA} shows that the baseline method generates only 10\% valid samples. Although it has the best Betti-0 performance compared to other methods, many connections are invalid and anatomically implausible (Fig. \ref{fig:qual}). Construct uses a tree constraint and generates 43 \% valid samples. Ours (k=0) strategy produces semantically valid samples, but has worse distribution statistics. In contrast, our method produces 100\% semantically valid samples with better adherence to the data distribution. Importantly, our model shows emerging link prediction capabilities on 30 test samples, which are not seen during training, as shown in Tab. \ref{tab:atm_labeler} and Fig. \ref{fig:downstream}. 

\paragraph{\textbf{Minimal Intervention During Sampling:}} In our experiments, the projection affects only ~2\% of all the generated edges. This limited intervention acts as a safeguard against error accumulation during sampling, rather than consistently overwriting the model's predictions. Hence, such minimal intervention sets our approach apart from largely post-hoc corrected random graph generators.

\paragraph{\textbf{Limitations:}} While we propose the first semantically consistent biological graph generator, our work considers solely the edge-deletion invariants and not the edge-insertion one. Hence, it can not guarantee constraints such as generating a single connected component. Additionally, since our point cloud generator is fixed during edge denoising, we can not refine any node coordinates, thus limiting generation flexibility. We aim to address these limitations in future work.